\documentclass{article}

\usepackage{microtype}
\usepackage{graphicx}
\usepackage{subfigure}
\usepackage{booktabs} 
\usepackage[table]{xcolor}
\usepackage{mathrsfs}

\usepackage{hyperref}

\usepackage[accepted]{icml2019}
\usepackage{todonotes}
\usepackage{amsfonts}       
\usepackage{nicefrac}       
\usepackage{microtype}      
\usepackage{wrapfig}
\usepackage{graphicx}
\usepackage{amsmath}
\usepackage{todonotes}

\icmltitlerunning{Humor in Word Embeddings: Cockamamie Gobbledegook for Nincompoops}

\begin{document}

\twocolumn[
\icmltitle{Humor in Word Embeddings:\\Cockamamie Gobbledegook for Nincompoops\\
\textcolor{red}{\normalsize{WARNING: This paper contains words that people rated humorous including many that are offensive in nature.}}}



\icmlsetsymbol{equal}{*}

\begin{icmlauthorlist}
\icmlauthor{Limor Gultchin}{ox}
\icmlauthor{Genevieve Patterson}{tr}
\icmlauthor{Nancy Baym}{ms}
\icmlauthor{Nathaniel Swinger}{lx}
\icmlauthor{Adam Tauman Kalai}{ms}
\end{icmlauthorlist}

\icmlaffiliation{ox}{University of Oxford}
\icmlaffiliation{ms}{Microsoft Research}
\icmlaffiliation{tr}{TRASH}
\icmlaffiliation{lx}{Lexington High School}

\icmlcorrespondingauthor{Limor Gultchin}{limor.gultchin@jesus.ox.ac.uk}

\icmlkeywords{Machine Learning, Humor, Word Embeddings}

\vskip 0.3in
]






\printAffiliationsAndNotice{}  

\begin{abstract}
While humor is often thought to be beyond the reach of Natural Language Processing, we show that several aspects of single-word humor correlate with simple linear directions in Word Embeddings. In particular: (a) the word vectors capture multiple aspects discussed in humor theories from various disciplines; (b) each individual's sense of humor can be represented by a vector, which can predict differences in people's senses of humor on new, unrated, words; and (c) upon clustering humor ratings of multiple demographic groups, different humor preferences emerge across the different groups. Humor ratings are taken from the work of Engelthaler and Hills (2017) as well as from an original crowdsourcing study of 120,000 words. Our dataset further includes annotations for the theoretically-motivated humor features we identify. 

\end{abstract}

\section{Introduction}

Detecting and generating humor is a notoriously difficult task for AI systems. While Natural Language Processing (NLP) is making impressive advances in many frontiers such as machine translation and question answering, NLP progress on humor has been slow. This reflects the fact that humans rarely agree upon what is humorous. Multiple types of humor exist, and numerous theories were developed to explain what makes something funny.
Recent research supporting the existence of single-word humor \cite{Engelthaler2017, nonwords} defines a more manageable scope to study with existing machine learning tools. Word Embeddings (WEs) have been shown to capture numerous properties of words \citep[e.g.,][]{mikolov2013efficient, mikolov2013linguistic}; coupled with single-word humor as a possible research direction, it is natural to study if and how WEs can capture this type of humor. To assess the ability of WEs to explain individual word humor, we draw on a long history of humor theories and put them to the test.


To many readers, it may not be apparent that individual words can be amusing in and of themselves, devoid of context. However, \citet{Engelthaler2017}, henceforth referred to as EH, found some words consistently rated as more humorous than others, through a crowdsourced study of about five thousand nouns. We first use their publicly available 5k mean word-humor ratings to identify a ``humor vector,'' i.e., a linear direction, in several WEs that correlate (over $0.7$) with these 5k mean humor ratings. While these correlations establish statistical significance, little insight is obtained into how the embeddings capture different aspects of humor and differences between people's senses of humor. 

To complete this picture, we performed crowdsourcing studies to create additional datasets which we make publicly available: (a) beginning with a set of 120k common words and phrases chosen from a word embedding, a crowdsourcing filtering process yielded a set of 8,120 and a further set of 216 words\footnote{These words, including {\em gobbledegook} and {\em nincompoops}, were rated as more humorous than the words in the EH study, which used common words from psychology experiments. The top-rated EH words were \textit{booty, tit, booby, hooter,} and \textit{nitwit}.} (in Appendix \ref{ap:funniest}) rated most humorous, (b) over 1,500 crowd workers rated these latter 216 words through six-way comparisons each yielding a personal first choice out of dozens of other personally highly-ranked words, and (c) 1,500 words (including highly-rated words drawn from these sets) were each annotated by multiple workers according to six humor features drawn from the aforementioned theories of humor. 

Our analysis suggests that individual-word humor indeed possesses many aspects of humor that have been discussed in general theories of humor, and that many of these aspects of humor are captured by WEs. For example, `incongruity theory,' which we discuss shortly, can be found in words which juxtapose surprising combinations of words like {\em hippo} and {\em campus} in {\em hippocampus}. Incongruity can also be found in words that have surprising combinations of letters or sounds (e.g. in words borrowed from foreign languages). The `superiority theory' can be clearly seen in insulting words such as {\em twerp}. The `relief theory' explains why humor may be found in taboo subjects and is found in subtle and not-so-subtle sexual and scatological connotations such as {\em cockamamie} and {\em nincompoop}. The most humorous words are found to exhibit multiple humor features, i.e., funny words are funny in multiple ways. We show that WEs capture these features to varying extents. 

Further, as is understood about humor in general, single-word humor is found to be highly subjective. We show here that individual senses of humor can be well-represented by vectors. In particular, an embedding for each person's sense of humor can be approximated by averaging vectors of a handful of words they rate as funny, and these vectors generalizes to predict preferences for future unrated words.  When clustered, these vector groups differ significantly by gender and age, and thus identify demographic differences in humor taste. We further introduce a ``know-your-audience'' test which shows that these sense-of-humor vectors can also be meaningfully differentiated using only a handful of words per person and generalize to make predictions on new unrated words.

While it would be easy to use Netflix-style collaborative filtering to predict humor ratings, WEs are shown to \textit{generalize}: given humor ratings on some words, vector representations are able to predict mean humor ratings, humor features and differences in sense of humor on test words that have no humor ratings. 

\smallskip\noindent
\textbf{Implications}.
Our study suggests that some of the fundamental features of humor are computationally {\em simple} (e.g., linear directions in WEs). This further suggests that the difficulty in understanding humor may lie in understanding rich domains such as language. If humor is simple, then humorous sentences may be hard for computers due to inadequate sentence embeddings: if you don't speak Greek, you won't understand a simple joke in Greek. Conversely, it also suggests that as language models improve, they may naturally represent humor (or at least some types of humor) without requiring significant innovation.

Furthermore, understanding single-word humor is a possible first step in understanding humor as it occurs in language. A natural next step would be to analyze humor is phrases and then short sentences, and indeed the 120,000 tokens rated include 45,353 multi-word tokens and the set of 216 tokens rated funniest includes 41 multi-word tokens.

\smallskip\noindent
\textbf{Organization}. Section \ref{sec:features} defines humor features drawn from humor theories and briefly discusses related work. Section \ref{sec:data} describes the word embeddings and data used. Section \ref{sec:predict} covers prediction of aggregate humor ratings from embedding vectors. Section \ref{sec:clustering} uses embedding representations to uncover group humor differences using unsupervised learning. Section \ref{sec:features predict} analyzes how word embeddings captures each of the theory-based humor features, while Section \ref{sec:diffs} analyzes how WEs capture differences in individual senses of humor.

\section{Relevant Features from Humor Theories}\label{sec:features}

In order to identify traces of single word humor in WEs, we first had to identify features of humor to look for. Defining humor, however, is a problem that was difficult enough to give food for thought to some of the most famous thinkers in history. Therefore, aware of our limitations, we do not seek to offer a definition of our own, but rather rely on the multiple existing theories from the social sciences, dating at least as far back as the philosophers of ancient Greece.
Plato, followed by Hobbes and Descartes, all contributed to a \textit{superiority theory} view of humor, which was formalized in the 20th century \cite{sep-humor}. However, since the 18th century, two new theories of humor surfaced and became much more widely acceptable: the \textit{relief theory} and the \textit{incongruity theory.}

The \textit{relief theory} offers a physiological explanation, which argues that laughter functions as a pressure relief for the nervous system, sometimes as a means to address taboo subjects that cause such stress. Proponents of this theory were Lord Shaftesbury \citeyearpar{cooper1709sensus}, Herbert Spencer  \citeyearpar{spencer1911essays} and Sigmund Freud \citeyearpar{freud1960jokes}. \textit{Incongruity theory} explains humor as a reaction to moments of uncertainty, in which two non-related issues which seem to be unfitting are juxtaposed in one sentence or event, but are then resolved. This theory could be seen as an explanation of the intuitive importance of punch lines in common jokes, where the ``set up'' builds an expectation that the ``punch line'' violates. Albeit confusing or disorienting logically, the revelation of the incongruity creates a humorous event when this contradiction violates an expectation, in a harmless, benevolent way. Among the supporters of this method were Kant \citeyearpar{kant1790critique}, Kierkegaard \citeyearpar{postscript1941trans}  and Schopenhauer \citeyearpar{schopenhauer1844arthur}.

These theories are also relevant to word humor. Based on these theories and other discussions of word humor \cite{hundredFunniestBeard,bown2015writing}, we consider the following six features of word humor:
\begin{enumerate}
\item Humorous {\em sound} (regardless of meaning): certain words such as {\em bugaboo} or {\em razzmatazz} are funny regardless of their meaning. This feature is related to the incongruity theory in that an unusual combination of sounds that normally do not go together can make a word sound funny. This feature is also consistent with the comedy intuition of Neil Simon, Monty Python, Dr. Seuss and the like, who discuss funny-sounding words and feature them in their works \citep[see, e.g.][]{bown2015writing}.
\item {\em Juxtaposition}/Unexpected incongruity: certain words are composed of terms which are otherwise completely unrelated. For example, there is little relation between the {\em hippocampus} part of the brain and the mostly unrelated words {\em hippo} and {\em campus}. This feature is motivated by incongruity theory.
\item {\em Sexual} connotation: some words are explicitly sexual (such as {\em sex}) or have sexual connotations (such as {\em thrust reverser}). The fact that some people find those words funny can be explained by the \textit{`relief theory'}, which suggests humor is a venting mechanism for social taboos. This was also discussed in the context of computational humor by Mihalcea et al. \cite{Mihalcea:2005:MCL:1220575.1220642}.
\item {\em Scatological} connotation: some words have connotations related to excrement to varying degrees, such as {\em nincompoop} or {\em apeshit}. The justification of this feature is similar to the sexual connotation feature above.
\item {\em Insulting} words: in the context of word humor, the superiority theory suggests that insulting words may be humorous to some people.
\item {\em Colloquial} words: extremely informal words such as {\em crapola} can be surprising and provide relief in part due to their unusually informal nature.
\end{enumerate}
We study the extents to which these features correlate with humor and how well WEs capture each one. Humor is known to vary by culture and gender, and EH analyzed age and gender differences in word humor. They also found strong correlations between humor ratings and word frequency and length, with shorter and less frequent words tending to be rated higher. This is interesting because word length and word frequency are strongly anti-correlated \cite{strauss2007word}. 

\subsection{Related Work in Computational Humor}\label{sec:comphumor}
Many of these features of humor have been considered in prior work on computational humor \citep[see, e.g.][]{Mihalcea:2005:MCL:1220575.1220642}, including early work such as HAHAcronym \cite{stock2003hahacronym}, which focused on producing humorous acronyms. More recently, \citet{van2018semeval} focused on humor detection, \citet{barbieri2014modelling} utilized Twitter for the automatic detection of irony and 
social media was further used to explore automatic classification of types of humor, such as anecdotes, fantasy, insult, irony, etc. \cite{raz2012automatic}. Other work studies visual humor \cite{chandrasekaran2016we} or humorous image captions \cite{shahaf2015inside, radev2015humor}. Other recent work has studied satire \cite{satire2019}.
WEs have been used as features in a number of NLP humor systems \citep[e.g.][]{chen2018humor, hossain2017filling, joshi2016word}, though the humor inherent in individual words was not studied.

\subsection{Other Work on Individual-Word Humor}
In addition to EH, another psychology study found consistent mean humor differences in {\em non-word} strings \cite{nonwords}. Outside of research, numerous lists of inherently funny words compiled by comedians and authors have been published \citep[e.g.][]{hundredFunniestBeard, tropes}. 

Finally, some comedians recommend humorous word choice. For instance, the character Willie in the Neil Simon play and subsequent film \textit{The Sunshine Boys} says:
\begin{quote}
Fifty-seven years in this business, you learn a few things. You know
what words are funny and which words are not funny. Alka Seltzer
is funny. You say ``Alka Seltzer" you get a laugh\ldots. Words with k in
them are funny. Casey Stengel, that's a funny name. Robert Taylor is
not funny. Cupcake is funny. Tomato is not funny. Cookie is funny.
Cucumber is funny. Car keys. Cleveland is funny. Maryland is not
funny. Then, there's chicken. Chicken is funny. Pickle is funny. All
with a k. Ls are not funny. Ms are not funny. \cite{simon1974sunshine}  \citep[as cited in][]{kortum2013funny}.
\end{quote}
This suggests that understanding funny words in isolation may be helpful as a feature in identifying or generating longer humorous texts.

\section{Data}\label{sec:data}
As is common, we will abuse terminology and refer to both the words and multi-word tokens from the embeddings as words, for brevity. When we wish to distinguish one word from multiple words, we write {\em single words} or {\em phrases}. 

\subsection{Embeddings}
WEs fit each word to a $d$-dimensional Euclidean vector based on text corpora of hundreds of billions of words. We consider several publicly-available pre-trained embeddings all using $d=300$ dimensions, trained on news and web data using different algorithms. The first embedding we consider, referred to as GNEWS, is the well-studied WE trained using the word2vec algorithm on about 100 billion tokens from Google News~\cite{mikolov2013efficient}. The second two embeddings, referred to as WebSubword and WebFast, were both trained using the fastText algorithm on a web crawl of 600 billion tokens~\cite{mikolov2018advances}. The difference is that WebSubword is trained using subword information and can therefore be evaluated on virtually any word or phrase, whereas WebFast like every other model in this paper can only be evaluated on words in its vocabulary. Finally, we consider an embedding trained on a similar web crawl of 840 billion tokens using the gloVe algorithm, referred to as WebGlove~\cite{pennington2014glove}.


\subsection{The Engelthaler-Hill Dataset}
Our first source of data is the EH dataset, which is publicly available \cite{Engelthaler2017}. It provides mean ratings on 4,997 single words, each rated by approximately 35 raters on a scale of 1-5. They further break down the means by gender (binary, M and F) and age (over 32 and under 32). However, since the EH data is in aggregate form, it is not possible to study questions of individual humor senses beyond age and gender. 

\subsection{Our Crowdsourcing Studies}
The EH data serves the purpose of finding a humor direction correlating with mean humor ratings. However, in order to better understand the differences between people's preferences among the funniest of words, we sought out a smaller set of more humorous words that could be labeled by more people. Eligible words were lower-case words or phrases, i.e., strings of the twenty-six lower-case Latin letters with at most one underscore representing a space. In our study, we omitted strings that contained digits or other non-alphabetic characters. 

English-speaking labelers were recruited on Amazon's Mechanical Turk platform. All labelers identified themselves as fluent in English, and 98\% of them identified English as their native language. All workers were U.S.-based, except for the results in Table \ref{table:nationality} of the appendix. We study a subset of 120,000 words and phrases from GNEWS, chosen to be the most frequent alphabetic lower-case entries from the embedding. While our study included both words and phrases, for brevity, in the tables in this paper we often present only single words. The list of 120,000 strings is included in the public dataset we are making available alongside this paper.\footnote{\url{https://github.com/limorigu/Cockamamie-Gobbledegook}}

\smallskip\noindent\textbf{Humor rating study.}
Our series of humor rating studies culminated in a set of 216 words with high mean humor ratings, which were judged by 1,678 U.S.-based raters. In each study, each participant was presented with random sets of six words and asked to select the one they found most humorous. In the first study only, the interface also enabled an individual to indicate that they found none of the six words humorous. We treat the selected word, if any, as being labeled positive and the words not selected as being labeled negative. Prior work on rating the humor in non-words found similar results for a forced-choice design and a design in which participants rated individual words on a Likert-scale \cite{nonwords}. 
To prevent fatigue, workers were shown words in daily-limited batches of up to fifty sextuple comparisons over the course of weeks. No worker labeled more than 16 batches. 

We refer to the three humor-judging studies by the numbers of words used: 120k, 8k, and 216. In the 120k study, each string was shown to at least three different participants in three different sextuples. 80,062 strings were not selected by any participant, consistent with EH's finding that the vast majority of words are not found to be funny. The 8k study applied the same procedure (except without a ``none are humorous'' option) to the 8,120 words that were chosen as the most humorous in a majority of the sextuples they were shown. Each word was shown to at least 15 different participants in random sextuples. To filter down to 216 words, we selected the words with the greatest fraction of positive labels. However, several near duplicate words appeared in this list. To avoid asking people to compare such similar words in the third stage, amongst words with the same stem, the word with the greatest score was selected. For instance, among the four words {\em wank}, {\em wanker}, {\em wankers}, and {\em wanking}, only {\em wanker} was selected for the third round as it had the greatest fraction of positive votes. The 216 words are shown in Appendix \ref{ap:funniest}.

In the 216 study, each participant selected not only a set of 36 ``humorous'' words, but further sets of 6 ``more humorous'' words and a single most humorous word, as follows. The 216 words were first presented randomly in 36 sextuples comprising the 216 words. In the same task, the 36 chosen words from these sextuples were shown randomly in 6 sextuples. The selected words were presented in a final sextuple from which they selected a single word. We associate a rating of 3 with the single funniest word selected at the end, a rating of 2 with the five other words shown in the final sextuple, a rating of 1 with the thirty words selected that did not receive a rating of 2 or 3, and a rating of 0 with the 180 unselected words. This process was chosen for efficiency as it requires only 43 clicks per user to identify three tiers of words that are rated highly.  However, the randomness in the initial permutation of each user's words introduces noise into the ratings. Nonetheless, it is easy to see that if a participant answers consistent with a secret internal ordering $\pi$, the expected values of the ratings (0-3, as defined above) of words would also be ordered according to $\pi$.



\noindent\textbf{Feature annotation study.}\label{sec:featureAnnotationStudyDescription}
The feature annotation study drew on the same worker pool. 1,500 words were chosen from the original 120,000 words by including the 216 words discussed above plus random words (half from the 8k study and half from the 120k). We asked participants to annotate six features discussed earlier, namely {\em humorous sound, juxtaposition, colloquial, insulting, sexual connotation, and scatological connotation}. Each feature was given a binary ``yes/no'' annotation, and results were averaged over at least 8 different raters per feature/word pair.

In each task, each rater was given the definition of a single feature and was presented with 30 words. They were asked to mark all words which possessed that feature by checking a box next to each word. A small number of workers were eliminated due to poor performance on gold-standard words. Further experimental details are in Appendix \ref{ap:further-experiment-details}. 

\section{Identifying a Mean Humor Direction}\label{sec:predict}
In prior work, the vector difference between embeddings of word pairs such as {\em trees:tree} and {\em she:he} was shown to be highly consistent with human ratings of concepts such as noun plurality or (binary) gender. While it is difficult to identify a word pair that represents humor, a linear direction (i.e., vector) in the embedding may still correlate strongly with mean humor ratings. 

Hence, we begin by fitting a simple least-squares linear regression model, predicting the EH mean humor rating for each word in the dataset from its 300-dimensional embedding. The resulting ``humor vector'' is of course not humorous as a collection of 300 numbers, but its inner-product with other word vectors is found to correlate significantly with humor. Table \ref{tab:EHcorrelations} shows the Pearson correlation coefficients of the predicted humor ratings with the actual EH humor ratings on the 4,997 EH words. The first column uses all EH words for fitting the linear model and for evaluation but is prone to overfitting since the models have 300 dimensions. For the second column, 10-fold cross validation is used to predict the labels (so each fold is predicted separately as hold-out) and then the correlation between the predictions and EH ratings is computed. Finally, the process is repeated 100 times to compute mean correlations and 95\% confidence intervals.   

\begin{table}[h!]
\centering
\begin{tabular}{lll}\toprule
Embedding name & Corr. & Corr. (using CV)\\\midrule
GNEWS (word2vec)       & 0.721 & 0.675 $\pm$ 0.003\\
WebSubword (fastText)  & 0.767 & 0.729 $\pm$ 0.002\\
WebFast (fastText)         & 0.767 & 0.730 $\pm$ 0.002\\
WebGlove (gloVe)       & 0.768 & 0.730 $\pm$ 0.002\\
\bottomrule
\end{tabular}
 \caption{\label{tab:EHcorrelations} Correlation coefficients compared across embeddings using least-squares regression to predict EH humor ratings. Correlations with 10-fold cross-validation and confidence intervals also reported.}
\end{table}

\begin{figure}[h!]
\centering
    \includegraphics[width=0.45\textwidth]{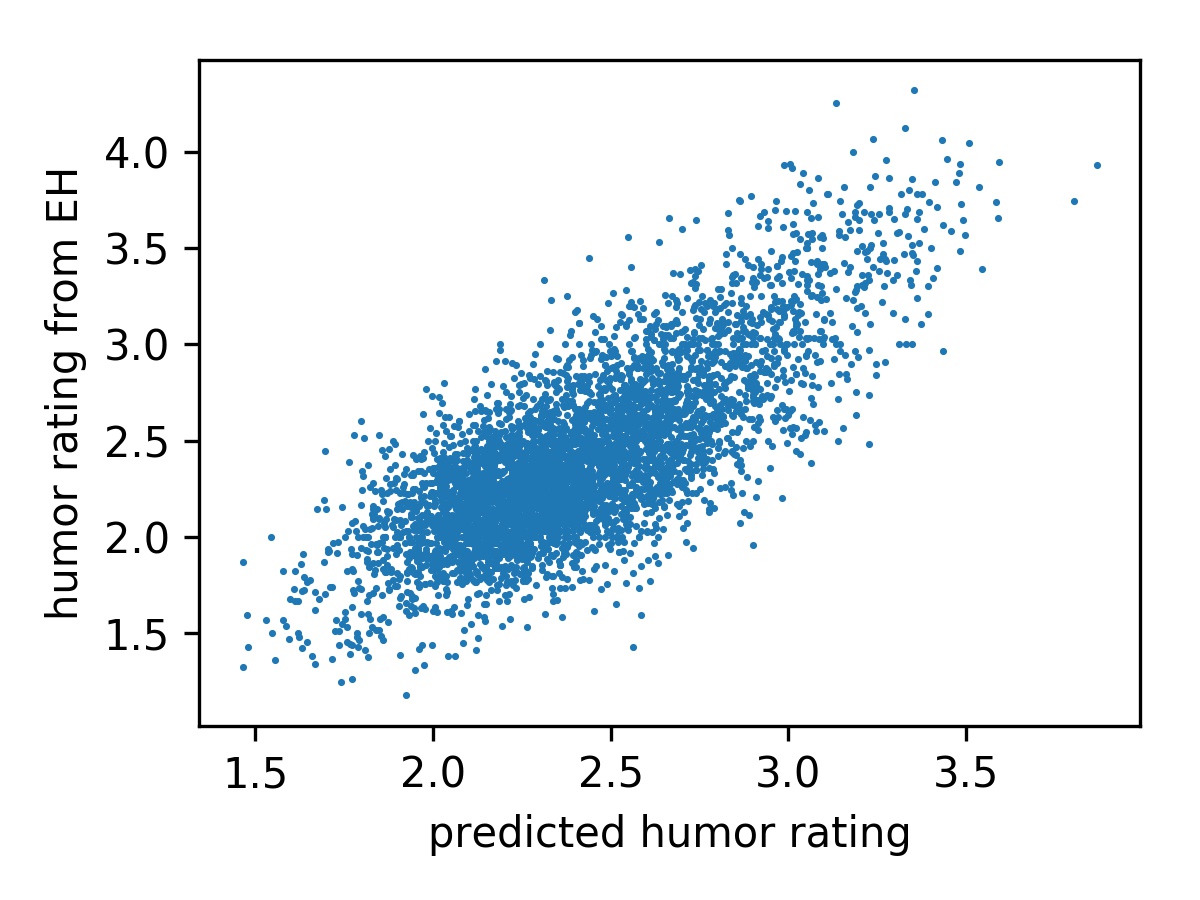}
  \caption{Humor rating predictions from a linear regression model trained on the WebSubword embedding vectors of words from the EH dataset, plotted against the EH mean humor ratings. See Table \ref{tab:EHcorrelations} for comparison with other embeddings.}
\end{figure}

We find that the vectors trained on the Web crawls slightly outperform GNEWS. This may be because of improved training algorithms but more likely the fact that they all use at least sixfold more training data. (We also note that a number of additional fastText and gloVe embeddings, trained on different and generally smaller corpora, were found to have smaller humor rating correlations, presumably due to the fewer tokens for training.) Also consistent with prior work \cite{bojanowski2017enriching}, the addition of subword information does not significantly change the performance of the fastText model. However, it does conveniently make it possible to compute embeddings of all words and phrases and thus compare it to the GNEWS embedding on the larger dataset (see Section \ref{sec:features predict}). 

Although the set of 5k words used in this evaluation is somewhat limited, it is a verified dataset used in a previous study of single word humor. Additionally, since the words are common to all the embeddings, it allows us to draw a comparison across embeddings. As we will show later, in the broader data collection task we preformed we found that many of the words which were rated as funniest were not included in EH and are less commonly used. Nonetheless, linear directions in the word embeddings correlated with the differences in humor ratings across words.

\begin{table*}[h]
\caption{Raters were clustered based on their sense-of-humor embeddings, i.e., the average vector of the words they rated funny. For each cluster, the single words that differentiate each rating cluster are shown together with scores on six humor features and demographics. Note that the demographics differences emerged from clustering the sense-of-humor vectors---demographics were not used during clustering.\label{tab:clusters}\smallskip
}

\centering

\begin{tabular}{llllll}\toprule
            & Cluster 1          & Cluster 2          & Cluster 3          & Cluster 4          & Cluster 5         \\\midrule
word 1      &  gobbledegook       & tootsies           & clusterfuck        & whakapapa          & dickheads         \\
word 2      &  kerfuffle          & fufu               & batshit            & codswallop         & twat              \\
word 3      &  hullaballoo        & squeegee           & crapola            & dabbawalas         & cocking           \\
word 4      &  razzmatazz         & doohickey          & apeshit            & pooja              & titties           \\
word 5      &  gazumped           & weenies            & fugly              & spermatogenesis    & asshattery        \\
word 6      &  boondoggle         & muumuu             & wanker             & diktats            & nutted            \\
word 7      &  galumphing         & thingies           & schmuck            & annulus            & dong              \\
word 8      &  skedaddle          & wigwams            & arseholes          & chokecherry        & wanker            \\
word 9      &  guffawing          & weaner             & dickheads          & piggeries          & cockling          \\
word 10      &  bamboozle          & peewee             & douchebaggery      & viagogo            & pussyfooting      \\
\midrule
sound       & \textbf{1.11} &                1.02 &                0.97 &                1.02 &                0.90\\
scatological&                0.80 &                0.99 & \textbf{1.15} &                0.89 &                1.14\\
colloqial   &                0.95 &                1.00 & \textbf{1.14} &                0.87 &                1.02\\
insults     &                0.86 &                0.90 & \textbf{1.23} &                0.84 &                1.12\\
juxtaposition&                0.89 &                0.86 &                0.99 &                1.10 & \textbf{1.13}\\
sexual      &                0.81 &                0.91 &                0.99 &                1.00 & \textbf{1.25}\\
\midrule
female \%   &  70.3\%$^*$            &  57.5\%            &  53.8\%            &  52.4\%            &  35.2\%$^*$           \\
mean age    &  38.6                &  37.4                &  42.3$^*$                &  37.2                &  34.7$^*$               \\
\bottomrule
\end{tabular}
\\ \smallskip
\textbf{Bolded numbers} represent the highest mean value obtained for that feature in any of the clusters. \\
$^*$statistically significant difference with p-value $< 10^{-6}$.
\end{table*}

While we see statistically significant correlations between humor vectors and EH ratings in all evaluated embeddings, several questions remain. First, do WEs capture the rich structure in the differences between individuals' senses of humor?
This cannot be answered from the EH data as it only reports aggregate mean ratings, not to mention that it does not contain many of the funniest words. A second and related question is, what is the nature of this correlation? For example, it could be the case that only one topic such as words with sexual connotations are both rated as highly humorous and captured by a linear direction in the word embedding, perhaps due to the fact that they often appear in similar contexts. To what extent are more subtle features such as humorous sounding words also captured? These questions motivated the collection of our crowdsourced datasets, focusing on differences in individual-based humor ratings and humor feature annotations.

\section{Clustering Sense-of-Humor Embeddings}\label{sec:clustering}
It is well known that humor differs across groups, cultures, and individuals.  We hypothesize that an individual's sense of humor can be successfully embedded as a vector as well. More specifically, if each person's ``sense-of-humor'' embedding is taken to be the vector average of words they rate as funny, this may predict which new, unrated words different people would find funny. 

For exploratory purposes, we next clustered the raters based on their sense-of-humor embeddings, i.e., the normalized average GNEWS vector of the 36 words they rated positively. We used K-means++ in sci-kit learn with default parameters \cite{scikit-learn}. 

For each cluster $i$ and each word $w$, we define a relative mean $\mu_i(w)$ for that cluster to be the difference between the fraction of raters in that cluster who rated that word positively and the overall fraction of raters who rated that word positively. Table \ref{tab:clusters} shows, for each cluster, the ten single words with maximum relative mean for that cluster (phrases are not displayed to save space). 
Table \ref{tab:clusters} also shows, for each feature of humor, the mean feature score of words rated positively by the individuals in that cluster, normalized by the overall mean. The number of clusters was chosen to be $K=5$ using the elbow method \cite{elbow}, though different $K$ values, different embeddings, and different random seeds exhibit qualitatively similar clusters. 

The emergent demographic differences between the clusters that emerged are surprising, given that the clusters were formed using a small number of word ratings (not demographics). First, Cluster 1 is significantly female-skewed (compared to overall 53.4\% female), Cluster 5 is male-skewed and younger, and Cluster 3 is older. How statistically significant are the age and gender differences between the clusters? We estimate this by randomly shuffling and partitioning the users into groups with the same sizes as the clusters repeatedly $10^7$ times and computing statistics of the cluster means. From this, the 95\% confidence interval for mean age was $(36.9, 39.2)$ and for percentage female was  $(48.5, 58.2)$. Hence, the significant age differences were clusters 3 and 5 and the significant gender differences were in clusters 1 and 5. Moreover these four differences were greater than any observed in the $10^7$ samples.  

Table \ref{tab:clusters} also exhibits the mean feature ratings of the words rated funny by cluster members, normalized by the overall mean feature rating. Interesting differences emerged. Consistent with table \ref{table:gender-funny} (see appendix), strings in the female-dominated cluster (Cluster 1) were rated as more ``funny-sounding'' while strings in the male-dominated cluster (Cluster 5) scored higher on sexual connotation and juxtapositions. The ``older'' cluster 3 rated words with scatological connotations as more humorous as well as insults and informal words. While none of the humor features we mentioned stand out for clusters 2 or 4, the clustering suggests new features that may be worth examining in a further study. For instance, Cluster 4 appears to highly rate unfamiliar words that may be surprising merely by the fact that they are words at all. Cluster 2 seems to rate ``random'' nouns highly (see the analysis of {\em concreteness} in \citet{Engelthaler2017}) as well as words like ``doohickey'' and ``thingie'' which can be described as ``random" nouns.

Table \ref{table:gender-funny} (see appendix) reports the strings that males and females differed most on, sorted by confidence in the difference. A key difference between Tables \ref{tab:clusters} and \ref{table:gender-funny} is whether or not the WE was used --- in Table \ref{tab:clusters} the participants were clustered by their sense of humor embeddings without using their gender, while Table \ref{table:gender-funny} presents differences in word ratings without using the WE at all. It is interesting that the WE recovers gender with a nontrivial degree of accuracy from the sense-of-humor embeddings. In the next section, we discuss predicting humor features, many of which emerged naturally in the clusters, using word embedding vectors.

\section{Humor Features in the Embedding}\label{sec:features predict}
Recall that workers rated the six theoretically-grounded features of humor described in Section \ref{sec:features} on 1,500 words (including the funniest 216 word set), as described in section \ref{sec:featureAnnotationStudyDescription}. Using supervised learning, we compute each feature's ``predictability'' as follows. As in the previous section, we use least squares linear regression to fit a linear function to the feature values as a function of the 300-dimensional embeddings, and use 10-fold cross validation to form predictions of feature tagging on all words. As before, we then compute the correlation coefficient between the predictions and labels. This is repeated 100 times and the mean of the correlation coefficients is taken as the feature predictability. To compute feature correlation with humor, we use the mean ratings of the 216 words in the personalized data and assign a rating of -1 to any word not in that set. 

\begin{figure}
\includegraphics[width=3.2in]{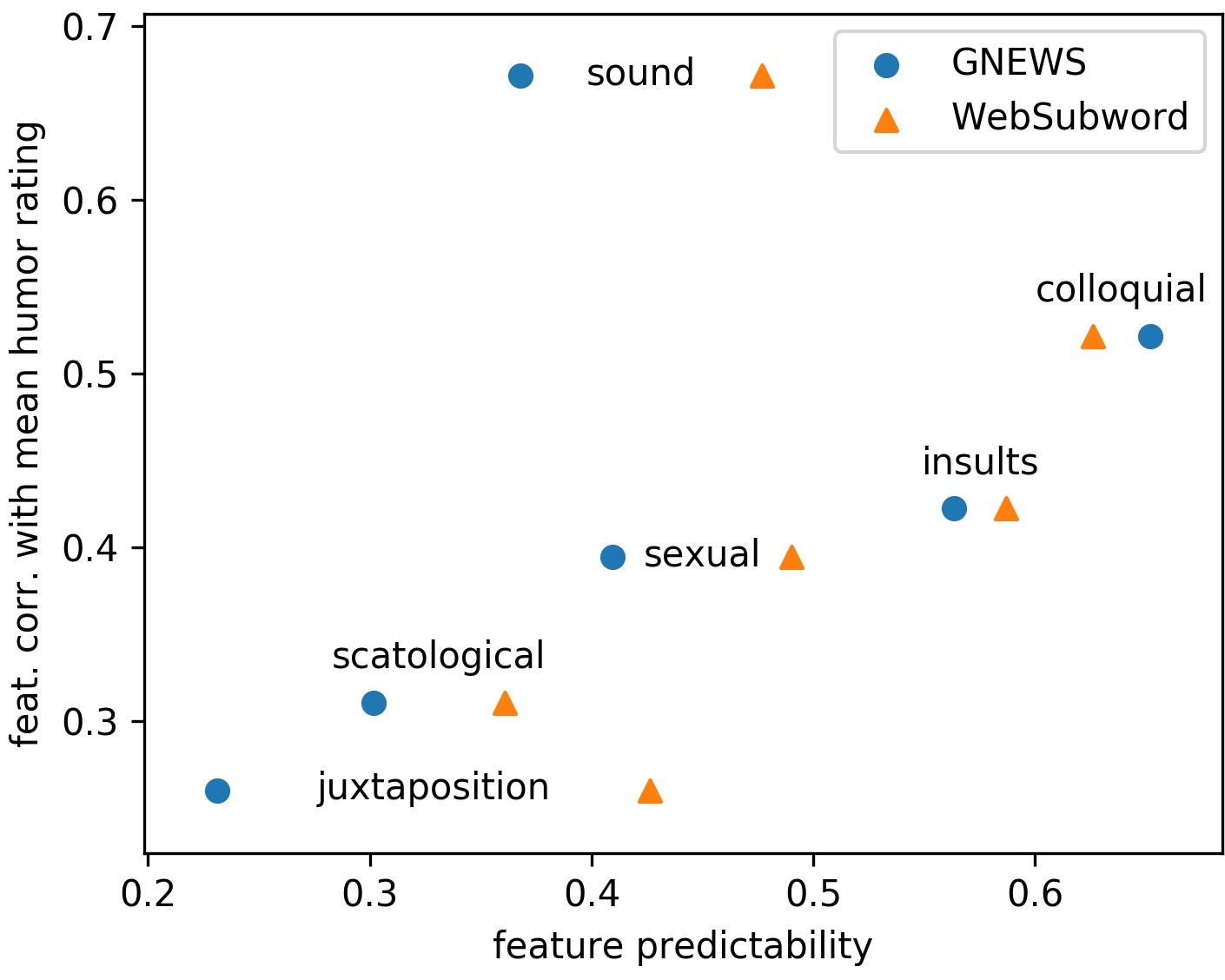}
\caption{Feature correlation with mean humor ratings vs. feature predictability (word-embedding correlations, based on GNEWS and WebSubword) for the six features on the 1,500 words with feature annotations.\label{fig:pred}}
\end{figure}

Figure \ref{fig:pred} shows these correlation with mean humor rating (y-axis) 
versus predictability (x-axis) across the words embeddings GNEWS and WebSubword. The GNEWS representation is well-suited for predicting colloquial (informal) words and insults, with correlations greater than 0.5, while the feature that was most difficult to predict with GNEWS was the juxtaposition feature, with a correlation slightly greater than 0.2. All the features had positive correlation with mean humor ratings, with funny sounding having the highest correlation. 

WebSubword predicted all features better than GNEWS with the exception of the colloquial feature. WebSubword yielded correlations which were equal or greater than 0.4 for five of the six identified humor features. The greatest difference was seen in the predictability of the juxtaposition feature, perhaps due to WebSubword's ability to extract meaningful information from subwords, highly relevant to a juxtaposed word which combines two shorter terms.

We can easily evaluate predictability and correlations with humor for an automatic feature such as word length. For word length, we find a predictability correlation coefficient of 0.518 indicating a good ability of a linear direction in the WE to predict word length, and a correlation with mean humor ratings of -.126, consistent with EH's findings that shorter words tend to be rated higher. 

\section{Predicting Individual Differences in Word Humor}\label{sec:diffs} 
We can use supervised learning to evaluate our earlier hypothesis that the average of the embeddings of the words a person finds funny captures their sense of humor in vector form. We consider the following test inspired by the comedy maxim, ``know your audience.'' In this test, we take two random people $p_1$ and $p_2$ with funniest words $w_1, w_2$ (rating of 3), respectively, where we require that $p_1$ rated $w_2$ as 0 and $p_2$ rated $w_1$ as 0. Note that this requirement is satisfied for 60\% of pairs of people in our data. In other words, 60\% of pairs of people had the property that neither one chose the other's funniest word even in the first round (of the 216 words study).  This highlights the fact that individual notions of humor indeed differ significantly. 

Given the two sets of 35 other words each participant rated positively, which we call the {\em training words}, and given words $\{w_1,w_2\}$ which we call the {\em test words}, the goal is to predict which participants rated $w_1$ or $w_2$ funniest. Just as a good comedian would choose which joke to tell to which audience, we use the WE to predict which person rated which test word as funny, based solely on the training words. For example, the training sets might be \{\textit{bamboozle, bejeezus, \ldots, wigwams, wingnuts}\} and \{\textit{batshit, boobies, \ldots, weaner, whakapapa}\} and test set to match might be the words \{\textit{poppycock, lollygag}\}. 

\begin{table}[h]
\caption{Success rates at know-your-audience tests, which test the ability of sense-of-humor embeddings to distinguish different raters based on sense of humor and predict which of two words each would find funny.\label{table:success}\smallskip
}
\centering
\begin{tabular}{lr}\toprule
Know-your-audience test & Success rate\\\midrule
Easy: disjoint sets of 35 training words. & 78.1\% \\
Normal: 35 training words.& 68.2\%\\
Hard: 5 training words. & 65.0\% \\\bottomrule
\end{tabular}
\end{table}

To test the embedding we simply average training word vectors and see which matches best to the test words. In particular, if $\vec{r}_1$ and $\vec{r}_2$ are the two training word vector averages and $\vec{v}_1$ and $\vec{v}_2$ are the corresponding test word vectors, we match $\vec{r}_1$ to $\vec{w}_1$ if and only if,
\begin{align*}
\|\vec{r}_1-\vec{v}_1\|^2+\|\vec{r}_2-\vec{v}_2\|^2 &< \|\vec{r}_1-\vec{v}_2\|^2+\|\vec{r}_2-\vec{v}_1\|^2\\
0 &< (\vec{r}_1-\vec{r}_2) \cdot (\vec{v}_1-\vec{v}_2)
\end{align*}
Simple algebra shows the above two inequalities are equivalent. Thus, the success rate on the test is the fraction of eligible pairs for which $(\vec{r}_1-\vec{r}_2) \cdot (\vec{v}_1-\vec{v}_2) > 0$. Note that this test is quite challenging as it involves prediction on two completely unrated words, the analog of what is referred to as a ``cold start'' in collaborative filtering (i.e., predicting ratings on a new unrated movie).

We also consider an easy and a hard version of this know-your-audience test. In the easy version, a pair of people is chosen who have disjoint training sets of positively rated words, indicating distinct senses of humors. In the hard version, we use as training words only the five words each person rated 2.
It is important to note that no ratings of the test words (or the other participants) are used, and hence it tests the WE's ability to generalize (as opposed to collaborative filtering). The test results given in Table \ref{table:success} were calculated by Brute force over all eligible pairs in the data (1,004 for the easy test and 818,790 for the normal and hard tests).



\section{Conclusions}
Can NLP-humor be closer than we think? This work suggests that many aspects of humor, including individual differences, may in fact be easier than expected to capture using machine learning -- in particular many correlate with linear directions in Word Embeddings.
In particular, each individual's sense-of-humor can be easily embedded using a handful of ratings, and that differences in these embeddings generalizes to predict different ratings on unrated words. We have shown that word humor possesses many features motivated by theories of humor more broadly, and that these features are represented WEs to varying degrees.



WEs are of course only one lens through which single word humor can be studied. Phonotactic, morphological and orthographic characteristics of humor can certainly be used to complement the WE approach. However, in this study, we show that WEs, popular NLP tools, are already enough to achieve meaningful progress in explaining single word humor. Furthermore, we offer our originally collected datasets, which we have made publicly available, as a useful contribution to future projects looking to complement and extend ours. 

There are numerous possible applications of word humor to natural language humor more generally. As discussed, comedians and writers are aware and indeed use such word choice to amplify humorous effects in their work. Our findings could therefore contribute to building aiding tools for creative writing tasks. Similarly, humorous word detection may help in identifying and generating humorous text. 
Moreover, our ratings could be used by text synthesis systems such as chat-bots which use WEs to tweak them towards or away from different types of humor (e.g., with or without sexual connotations), depending on the application at hand and training data. 


Finally, one approach to improving AI recognition and generation of humor is to start with humorous words, then move on to humorous phrases and sentences, and finally to humor in broader contexts. Our work may be viewed as a first step in this programme.

\textbf{Acknowledgements.} We would like to thank Kai-Wei Chang and anonymous reviewers for their helpful comments.

\bibliography{humor}
\bibliographystyle{icml2019}

\newpage
\appendix
\onecolumn

\section{Further Experiment Details}\label{ap:further-experiment-details}

We found many proper nouns and words that would normally be capitalized in the 120,000 most frequent lower-case words from GNEWS. To remove these words, for each capitalized entry, such as {\em New\_York}, we removed it and also less frequent entries (ignoring spacing) such {\em newyork} if the lower-case form was less frequent than the upper-case word according to the WE frequency. (For example, the WE has 13 entries that are equivalent to {\em New\_York} up to capitalization and punctuation.) 

To disincentivize people from randomly clicking, a slight pause (600 ms) was introduced between the presentation of each word and the interface required the person to wait until all six words had been presented before clicking. In each presentation, words were shuffled to randomize the effects of positional biases. The fractions of clicks on the different locations did suggest a slight positional bias with click percentages varying from 15.7\% to 18.6\%. 

We refer to the three humor-judging experiments by the numbers of words used: 120k, 8k, and 216. In the 120k experiment, each string was shown to at least three different participants in three different sextuplets. 80,062 strings were not selected by any participant, consistent with EH's finding that the vast majority of words are not found to be funny. The 8k experiment applied the same procedure (except without a ``none are humorous'' option) to the 8,120 words that were chosen as the most humorous in a majority (1/2 or more) of the sextuples they were shown. Each word was shown to at least 15 different participants in random sextuples. The list of 216 words is in the appendix. A slight inconsistency may be found in the published data in that we have removed duplicates where one person voted on the same word more than once, however in forming our sets of 8,120 and 216 words we did not remove duplicates.

\section{Further Materials and Findings}\label{ap:funniest}
The 216 top-rated words, sorted in order of mean humor ratings (funniest first), used in the main crowdsourcing experiment:

\noindent
asshattery, clusterfuck, douchebaggery, poppycock, craptacular, cockamamie, gobbledegook, nincompoops, wanker, kerfuffle, cockle pickers, pussyfooting, tiddlywinks, higgledy piggledy, kumquats, boondoggle, doohickey, annus horribilis, codswallop, shuttlecock, bejeezus, bamboozle, whakapapa, artsy fartsy, pooper scoopers, fugly, dunderheaded, dongles, didgeridoo, dickering, bacon butties, woolly buggers, pooch punt, twaddle, dabbawalas, goober, apeshit, nut butters, hoity toity, glockenspiel, diktats, mollycoddling, pussy willows, bupkis, tighty whities, nut flush, namby pamby, bugaboos, hullaballoo, hoo hah, crapola, jerkbaits, batshit, schnitzels, sexual napalm, arseholes, buffoonery, lollygag, weenies, twat, diddling, cockapoo, boob tube, galumphing, ramrodded, schlubby, poobahs, dickheads, fufu, nutjobs, skedaddle, crack whore, dingbat, bitch slap, razzmatazz, wazoo, schmuck, cock ups, boobies, cummerbunds, stinkbait, gazumped, moobs, bushwhacked, dong, pickleball, rat ass, bootlickers, skivvies, belly putter, spelunking, faffing, spermatogenesis, butt cheeks, blue tits, monkeypox, cuckolded, wingnuts, muffed punt, ballyhoo, niggly, cocksure, oompah, trillion dong, shiitake, cockling, schlocky, portaloos, pupusas, thrust reverser, pooja, schmaltzy, wet noodle, piggeries, weaner, chokecherry, tchotchkes, titties, doodad, troglodyte, nookie, annulus, poo poo, semen samples, nutted, foppish, muumuu, poundage, drunken yobs, yabbies, chub, butt whipping, noobs, ham fisted, pee pee, woo woo, squeegee, flabbergasted, yadda yadda, dangdut, coxless pairs, twerps, tootsies, big honkin, porgies, dangly, guffawing, wussies, thingies, bunkum, wedgie, kooky, knuckleheads, nuttin, mofo, fishmonger, thwack, teats, peewee, cocking, wigwams, red wigglers, priggish, hoopla, poo, twanged, snog, pissy, poofy, newshole, dugong, goop, whacking, viagogo, chuppah, fruitcakes, caboose, cockfights, hippocampus, vindaloo, holeshot, hoodoo, clickety clack, backhoes, loofah, skink, party poopers, civvies, quibble, whizzy, gigolo, bunged, whupping, weevil, spliffs, toonie, gobby, infarct, chuffed, gassy, crotches, chits, proggy, doncha, yodelling, snazzy, fusarium, bitty, warbled, guppies, noshes, dodgems, lard, meerkats, lambast, chawl

Tables \ref{table:gender-funny} and \ref{table:nationality} present binary comparisons of the funniest words by gender (female vs.~male) and nationality (India vs.~U.S.).

\begin{table*}[h!]
\caption{Among our set of 216 words (including phrases), the ten with most confident differences in ratings from people in India and the U.S.\  (using a two-sided t-test and Bonferoni correction). There is a strong (0.45) correlation between word length and difference in rating between U.S.~and India.\label{table:nationality}}
\centering
\begin{tabular}{llll}\toprule
Word rated funnier in India & adjusted p-value & Word rated funnier in U.S. & adjusted p-value\\\midrule 
poo poo & 6.2e-14 & codswallop & 4.5e-25\\
pissy & 2.4e-12 & craptacular & 5.7e-22\\
woo woo & 5.4e-12 & asshattery & 4.2e-20\\
poofy & 9.2e-11 & kerfuffle & 1.3e-19\\
gigolo & 3.2e-10 & gobbledegook & 3e-18\\
muumuu & 4.5e-10 & glockenspiel & 2.9e-17\\
pee pee & 6.2e-10 & clusterfuck & 2.4e-16\\
guppies & 2.4e-09 & ramrodded & 1.8e-13\\
gassy & 1e-07 & douchebaggery & 9.4e-13\\
boobies & 4.2e-07 & twaddle & 4.7e-12\\
\bottomrule\end{tabular}
\end{table*}

\begin{table*}[h!]
\caption{\label{tab:funniest}The ten words rated funniest in our study, their female/male mean rating discrepancy (if significant), and some features of these words.}
\centering
\rowcolors{2}{gray!25}{white}
\begin{tabular}{lccccccccc}\toprule
              & FM & sound     & juxtaposition& colloquial& insulting   & sexual     & scatological\\\midrule
asshattery    &   &            &  \checkmark & \checkmark & \checkmark  & \checkmark &  \checkmark\\
clusterfuck   & M &            &  \checkmark & \checkmark &             & \checkmark &  \checkmark\\
douchebaggery &   &            &             & \checkmark &  \checkmark &            & \\
poppycock     & M & \checkmark &  \checkmark &            &  \checkmark & \checkmark &\\
craptacular   &   &            &             &            &  \checkmark &            &  \checkmark\\
cockamamie    & F &            &             &            &  \checkmark & \checkmark & \\
gobbledegook  & F & \checkmark &             &            &             &            &  \\
nincompoops   & F & \checkmark &  \checkmark & \checkmark &  \checkmark &            & \checkmark\\
wanker        & M &            &             & \checkmark &  \checkmark & \checkmark & \\
kerfuffle     & F & \checkmark &             & \checkmark &             &            & \\\bottomrule
\end{tabular}
\end{table*}

\begin{table}[h]
\centering
\caption{\label{tab:annotations} 10 most highly annotated words in each humor feature category.}
\begin{tabular}{@{}llllll@{}}
\toprule
Colloquial & Insulting & Juxtaposition & Scatological & Sexual & Funny sounding \\ \midrule
dissing & fuckers & party poopers & dog poop & foreskins & lollygag \\
wee lad & douche & hippocampus & dung & scrotum & gobbledegook \\
clusterfuck & dickheads & pooch punt & pooper scoopers & nudism & ballyhoo \\
twat & nincompoops & port potties & poo poo & nutted & tiddlywinks \\
mofo & asshat & bacon butties & urination & blue tits & higgledy piggledy \\
fugly & wussies & sexual napalm & apeshit & boobies & schlubby \\
flimflam & twat & pickleball & poo & pussyfooting & hobnobbing \\
woo woo & crack whore & pussyfooting & butt cheeks & crotches & hoo hah \\
crack whore & rat ass & boob tube & crapola & vibrators & didgeridoo \\
nutjobs & smartass & jobholders rose & diaper clad & masturbating & poppycock \\ \bottomrule
\end{tabular}
\end{table}

\begin{table}[h!]
\caption{Among our set of 216 words (including phrases), the ten with most confident differences in ratings across gender (again using a two-sided t-test and Bonferoni correction).\label{table:gender-funny}}
\centering
\begin{tabular}{lrlr}\toprule
Word rated funnier by F & adjusted p-value & Word rated funnier by M & adjusted p-value\\\midrule 
whakapapa & 3.2e-04 & sexual napalm & 2.1e-11\\
doohickey & 0.0011 & poundage & 1.3e-05\\
namby pamby & 0.0014 & titties & 2.6e-05\\
hullaballoo & 0.003 & dong & 3.2e-05\\
higgledy piggledy & 0.0039 & jerkbaits & 7.4e-05\\
gobbledegook & 0.0047 & semen samples & 1.8e-04\\
schlocky & 0.008 & nutted & 0.0019\\
gazumped & 0.014 & cock ups & 0.0021\\
kooky & 0.026 & boobies & 0.0027\\
schmaltzy & 0.033 & nut butters & 0.004\\
 \bottomrule\end{tabular}
 \vspace{128in}
\end{table}

\end{document}